# Training the Convolutional Neural Network with Statistical Dependence of the Response on the Input Data Distortion

Igor M. Janiszewski[1], Vladimir V. Arlazarov[2,3,4], Dmitry G. Slugin[1,2]
[1]Federal Research Center Computer Science and Control of Russian Academy of Sciences, Moscow, Russia;
[2]Smart Engines Service LLC, Moscow, Russia;
[3]Institute for Information Transmission Problems of Russian Academy of Sciences, Moscow, Russia;
[4]Moscow Institute of Physics and Technology (State University), Moscow, Russia

## ABSTRACT

The paper proposes an approach to training a convolutional neural network using information on the level of distortion of input data. The learning process is modified with an additional layer, which is subsequently deleted, so the architecture of the original network does not change. As an example, the LeNet5 architecture network with training data based on the MNIST symbols and a distortion model as Gaussian blur with a variable level of distortion is considered. This approach does not have quality loss of the network and has a significant error-free zone in responses on the test data which is absent in the traditional approach to training. The responses are statistically dependent on the level of input image's distortions and there is a presence of a strong relationship between them.

## 1. INTRODUCTION

Nowadays the mobile devices market, in particular smartphones, grows explosively. Typical modern devices already have enough high-productive processors and high-quality cameras, which allows to solve many problems using neural networks directly on the device, without cloud technologies [1], [2]. One of these tasks is document recognition in a video stream, part of which is optical character recognition (OCR), that is typically performed using a classifying neural network. Convolutional neural networks are actively used in the task of character classification and recognition, as they provide a suitable balance between the complexity of the network (which affects the performance) and the recognition quality. The source data for such networks are symbols selected from the video stream. It is worth noting that the images of symbols are usually seriously different from the "ideal" in the original document due to the difference in lighting when shooting and glares from light sources, spatial distortions of the document on the frames, noise of the camera matrix, blurring the image at slow shutter speeds, shaking the device, and so on [3]. To improve the neural network classification, it is necessary not only to have a sufficient data set for training, but also to find a suitable augmentation model that can simulate various types of distortions and to supply the original set with missing examples [4], [5]. The result of the classifying neural network is a set of estimates of the symbol images, besides to the absolute values as a criterion for belonging to the class, these estimates can be interpreted as sort of "confidence" of the network in the classification. A certain threshold for estimates can be chosen empirically, below which the responses of the neural network are considered doubtful and require an additional analysis. Aside from the classification quality of the trained network (the percentage of correct answers on the test dataset), for practical use in real-world tasks, we would like to have the following features:

   a. The incorrect classifications of the network have a low confidence. It helps to avoid the most of incorrect classifications by choosing a suitable threshold.

   b. Monotonic dependence between confidence and the level of distortion of the original image. This is useful, for example, to integrate recognition results from many frames of video stream containing the same document, as it allows to make some assessment of the original image quality [6].

The paper proposes a method of learning convolutional neural network using the level of distortion in the construction of the extended network model. The original neural network has the same quality and obtains these properties.

## 2. METHOD DESCRIPTION

### 2.1 Problem Statement

Suppose we have an initial model $M$ of the convolutional network, which receives inputs $x_i, i \in 1 \ldots N$. The network response is a vector $s_i = M(x_i)$, $s_i \in R^K$, where $K$ – number of classes ($K > 1$). Let the activation function just before Softmax layer, which forms the logit vector $s_i$, is the linear rectifier ReLU6, so the outputs $z_i$ is in the interval [0;6]. Generally,

$$ReLUX(v) = \min(X, \max(0, v)) \quad (1)$$

Using vector $s_i$, it is possibly to predict the class $c_i = \underset{k}{\operatorname{argmax}}\, s_i^{(k)}$ to which the object $x_i$ belongs to and get the network response $s_i^{(k)}$ as a confidence score for each class, where

$$s_i^{(k)} = \frac{\exp(z_i^{(k)})}{\sum_{l=1}^{K} \exp(z_i^{(l)})} \quad (2)$$

The model $M$ is shown in Fig.1a.

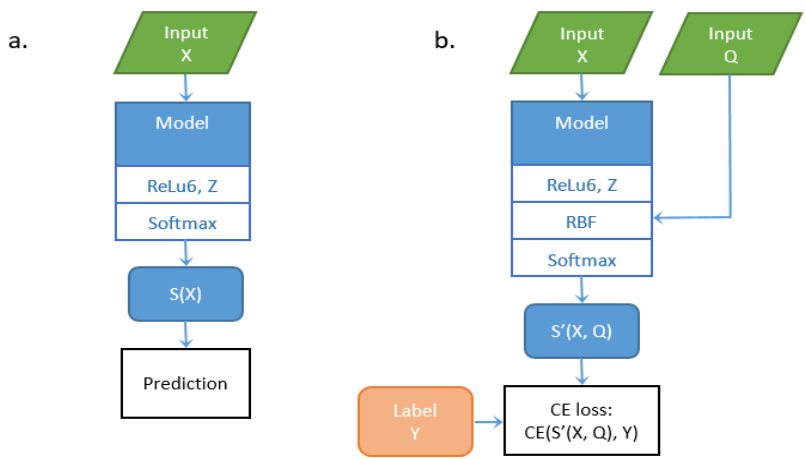

Figure 1. Networks $M$ and $M'$ architectures.

Assume that each object $x_i$ has some property $Q$ with a value of $q_i \in [a, b] \subset R^1$. Note that different objects can have the same property $q$. The task is to create and train a neural network $M'$, which contains the original network $M$ and has the property: for objects $x_i$ with values $q_1, q_2 \in [a, b]$ and $q_1 < q_2$ follows (3) where $Md$ is a median.

$$Md[s_i^{(c_i)}|q_i = q_1] > Md[s_j^{(c_j)}|q_j = q_2], \quad (3)$$

$M'$ network architecture is shown in Fig.1b and consists of a network $M$ with an added input layer of properties $Q$ and an additional intermediate RBF layer with a parametric activation function depending on Q. The answer of the network $M'$ is a logit vector $s_i' = M'(x_i, q_i)$. Let us consider in detail, step by step, the construction of $M'$.

### 2.2 Construction of Network

Step 1. Select values $p_{max}, p_{min} \in (0; 1.0)$ where $p_{max} > p_{min}$. Let's call these values "target values" for $q = a$ and $q = b$. For each $q_i \in (a, b)$ corresponded $p_i$ can be calculated using linear interpolation (4).

$$p_i = p_{max} - (p_{max} - p_{min})\frac{q_i - a}{b - a} \quad (4)$$

So, every $x_i$ with property $q_i$ has $p_i$ inside $[p_{min}; p_{max}]$. This value can be interpreted as target estimate of network $M$ for object $x_i$.

Step 2. Using logit function $p$

$$p(\zeta) = \frac{\exp(\zeta)}{\exp(\zeta)+(K-1)} \qquad (5)$$

and selected $p_i$ it is possible to obtain $\zeta_i$ with inverse logit function

$$\zeta_i = \ln(p_i \cdot (K-1)) - \ln(1-p_i) \qquad (6)$$

It is easy to see that the expression (5) can be derived from (2) under the assumption that $z_i^{(k)} = 0, k \neq c_i$. If response $z_i^{(c_i)}$ tends to $\zeta_i$ the confidence score $s_i^{(c_i)}$ will be close to target estimate $p_i$ in case that other $z^{(k)}$ are around zero.

Step 3. Introduce element-wise transformation on RBF layer where Gaussian function [7] is used as a radial-basis function:

$$G(z_i; A, \zeta_i, \sigma)^{(k)} = A \cdot \exp\left(-\frac{\left(z_i^{(k)}-\zeta_i\right)^2}{2\sigma^2}\right) \qquad (7)$$

Where $A$ is a height of peak ($A > 0$), $\zeta_i$ is a center of peak and $\sigma$ is a standard deviation ($\sigma > 0$).

Step 4. Categorical Cross Entropy Loss is used as a loss function (8) where $y_i^{(k)} = 1$ for correct class $k$ and 0 otherwise.

$$L = -\sum_{i=1}^{N}\sum_{k=1}^{K} y_i^{(k)} \cdot \log\left(s'^{(k)}_i\right) \qquad (8)$$

Thus, the constructed model $M'$ in learning process approximates the value of the vector $\{z_i\}$, corresponding to the correct class, to the center of the peak, as a result the confidence score of the model $M$ falls into the neighborhood of the target estimate for the corresponding $q_i$. The remaining values of $\{z_i\}$ are initially small and according to the domain of definition do not go beyond the neighborhood of zero because of the monotonic decrease of the Gaussian function from the center of the peak. When the training is completed, the RBF layer is removed from the model $M'$ and the network of the original $M$ architecture is obtained.

## 3. DATA MODEL

Well-known dataset of handwritten numbers MNIST [8] is used as a source data. As a model of data distortion, the Gaussian blur is considered [9]. Let's put the property $q = \sigma$ where $\sigma$ is a standard deviation - parameter of Gaussian blur. The samples of image's distortion in shown in Fig.2. The numbers of the training and testing datasets were increased using this augmentation in 10 times uniformly in interval $q \in [0; 4]$. As a result, size of train dataset is equal 600 000, the test is 100 000. It should be noted that for small values of $q$ the distortion is practically absent as the weights of neighboring pixels are rather small (see Fig.3). Because of that only samples with $q \geq 0.5$ was used for training.

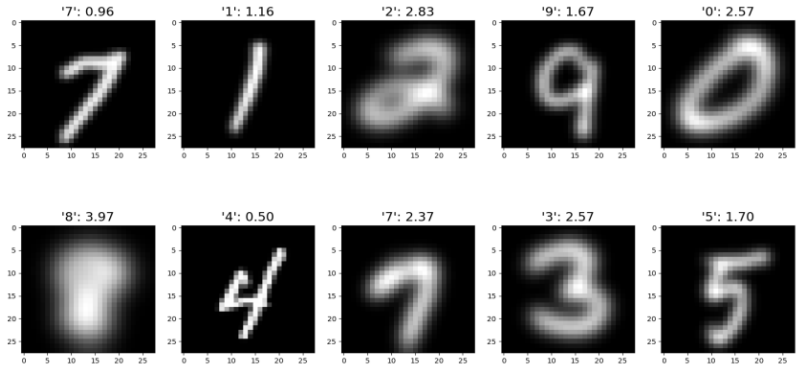

Figure 2. Samples of data augmentation depending on distortion parameter.

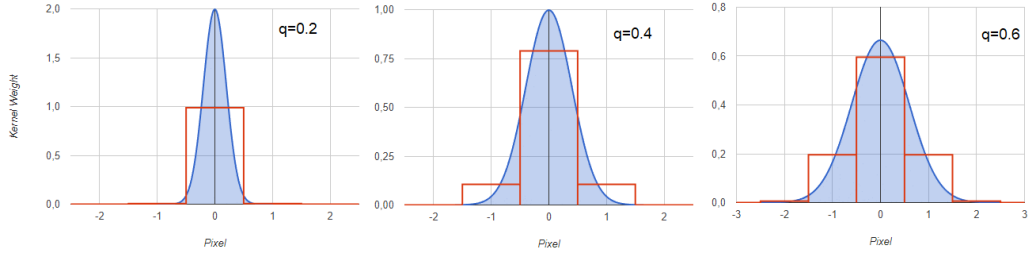

Figure 3. Weights of neighboring pixels depending on distortion parameter.

## 4. EXPERIMENT RESULTS

Consider the convolutional neural network like LeNet5 [10] with training on the selected data model. It consists of a combination of the following layers: a convolution, full connection and subsampling. A linear rectifier ReLU is used as an activation function, on the last layer there is a rectifier with a limit of the maximum value ReLU6. The visualization of the network architecture is shown in Fig.4.

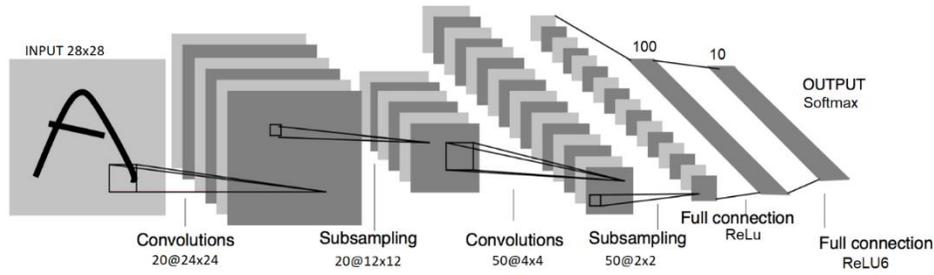

Figure 4. LeNet5 convolution network.

The training parameters were the following:

- number of classes $K = 10$;
- number of trainable weights 46 680;
- distortion $q \in [0; 4]$;
- $p_{max} = 0.6$, $p_{min} = 0.3$;
- peak $A = 6$;
- deviation $\sigma = 0.7$.

The training was performed using Keras with a limit of 200 epochs. The results on the test dataset are summarized in Table 1, where $M_1$ are the results of the network trained as is and $M_2$ as part of the $M'$ model.

Table 1. Experiment's results

| Network | Quality, % | Error-free rate, % | Spearman's correlation coefficient |
|---|---|---|---|
| $M_1$ | 99.08 | 0 | -0.12 |
| $M_2$ | 99.10 | 34.14 | -0.86 |

The accuracy of the trained networks $M_1$ and $M_2$ is almost equal within the stochasticity of the training process. Consider the distribution of confidences for correct and incorrect classifications depending on the level of distortion in the form of a correlation field. Fig.5 presents the field for the $M_1$ network, Fig.6 – for $M_2$.

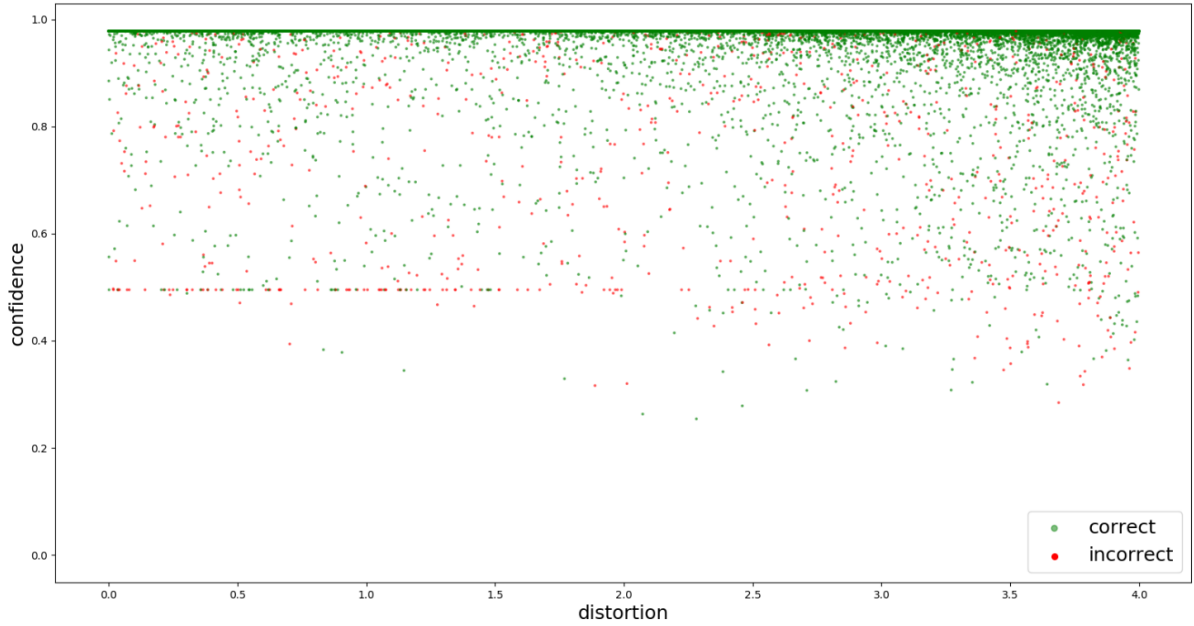

Figure 5. Correlation field for $M_1$.

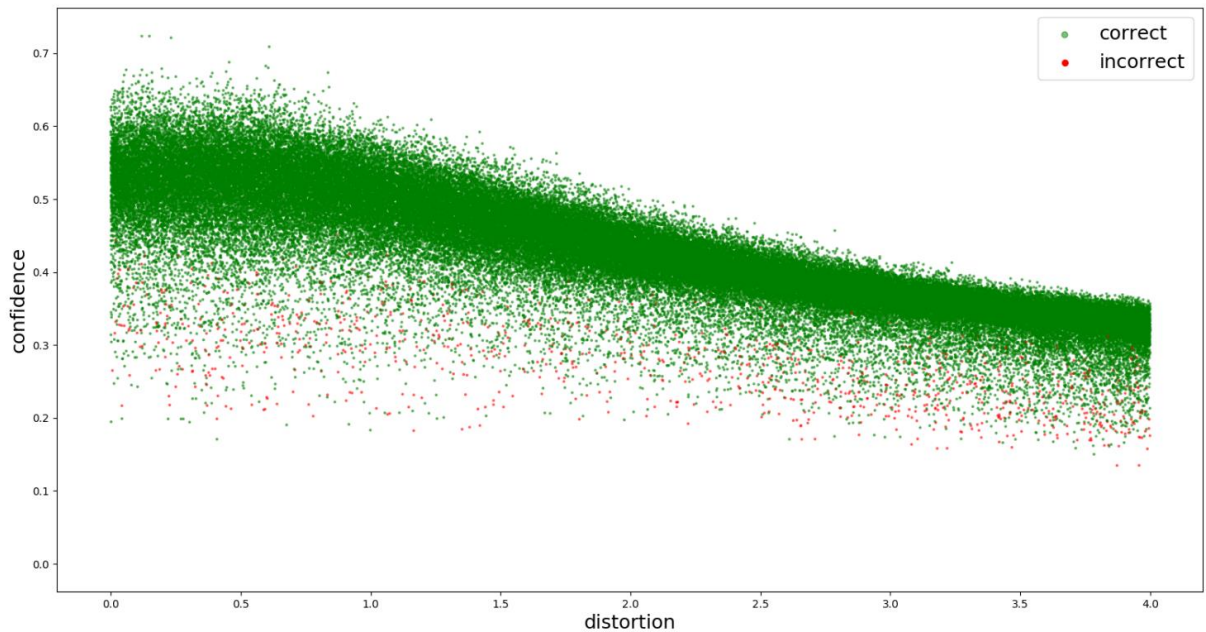

Figure 6. Correlation field for $M_2$.

For test dataset the error-free threshold — the maximum value of the confidence score with incorrect classifications and the error-free rate — the percentage of correct classifications with confidence above the error-free threshold were calculated. These values characterize the level of trust in the network responses and show the presence or absence of incorrect classifications with an abnormally high level of confidence. From Table 1 it follows that $M_2$ has a significant level of error-free rate while $M_1$ does not have it at all. In contrast to $M_1$, a correlation between the level of distortion and the confidence score is visually observed for the network $M_2$. To check this, the Spearman's correlation coefficient [11] for each network was calculated. The correlation coefficient for $M_1$ is small, while for $M_2$ it approaches to -1 that means a strong decreasing monotonic relationship between distortion and confidence.

## 5. REGRESSION ANALYSIS

The Spearman's correlation coefficient is only a characteristic indicating the possibility of a statistical relationship between the level of distortion and network confidence. To find the form of such dependence, a regression analysis was performed. For test dataset, the normality of residuals requirement, which is necessary for applying classical regression models, is not fulfilled. Therefore, we use quantile regression [12] to check a linear dependence of the network response on the level of distortion in terms of quantiles. This type of regression refers to nonparametric research methods, that is, which do not require a priori knowledge about the distribution of random variables. An additional feature of quantile regression is its outlier resistance.

Let's build a dependence model with a quantile equal to $\tau = 0.5$, i.e., the median regression. This means that at this distortion level, the network responses are equally likely to be greater or less than the value of the regression model. It is possible to build the models for different quantiles and then calculate the confidence interval of network responses for a given probability.

The quantile regression model is as follows

$$Q_\tau(y_i) = \beta_0(\tau) + \sum_{j=1}^{L} \beta_j(\tau) x_{i,j}, \quad i \in 1 \ldots N \tag{9}$$

Where $Q_\tau$ is the quantile of the level $\tau$, $x_j$ are the regressors (model factors), $L$ is the number of regressors. The values $\beta_k(\tau)$ are found by solving the minimization problem

$$\min_{\beta_0(\tau),\ldots,\beta_L(\tau)} \sum_{i=1}^{N} \rho_\tau \left( y_i - \beta_0(\tau) - \sum_{j=1}^{L} \beta_j(\tau) x_{i,j} \right) \tag{10}$$

Where loss-function $\rho_\tau$ defined as

$$\rho_\tau(x) = \tau \max(x, 0) + (1 - \tau) \max(-x, 0) \tag{11}$$

In our case, $L = 1$, $x_{i,j} = q_i$. From Fig.6 it is obvious that there is no global linear regression model, therefore, the entire interval of values $q \in [0; 4]$ was divided into some sub-intervals and built a model for each of them. Fig.7 shows quantile regression lines for each interval, as well as separately calculated real median values.

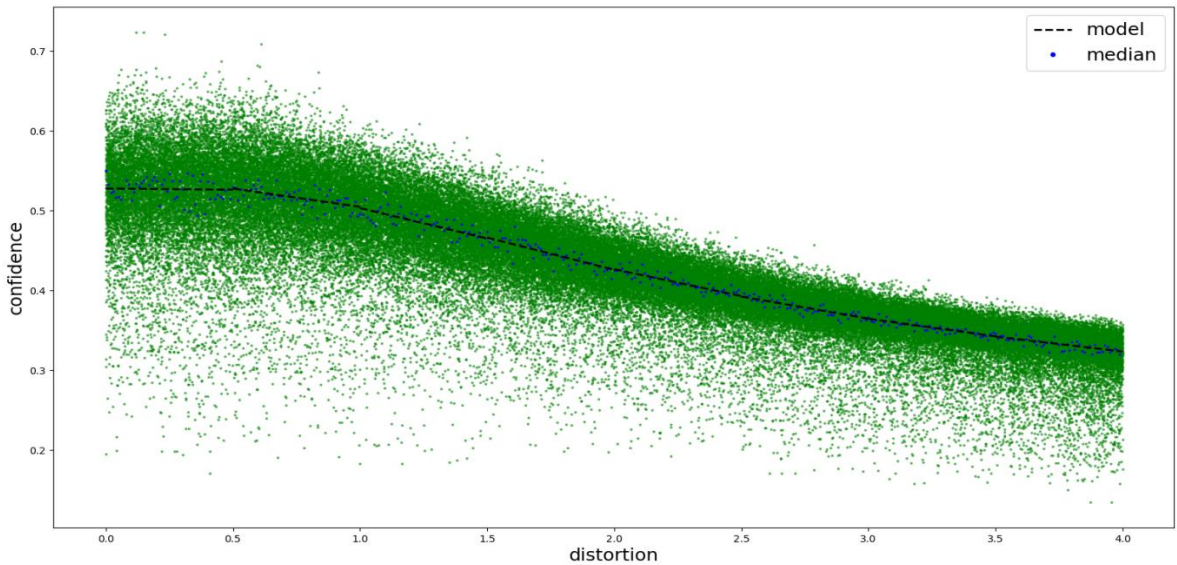

Figure 7. Interval model of median regression.

To check the adequacy of the constructed models for each interval, a goodness-of-fit test, based on the CUSUM process of the gradient vector [13], using GOFTest function of the QTools package for R was made. As can be seen from Table 2, the hypothesis of linearity of medians for 7 out of 8 intervals will not be rejected with significance level not less than 0.05.

Table 2. Calculated p-values depending on interval

| Interval | [0;0.5] | [0.5;1.0] | [1.0;1.5] | [1.5;2.0] | [2.0;2.5] | [2.5;3.0] | [3.0;3.5] | [3.5;4.0] |
|---|---|---|---|---|---|---|---|---|
| p-value | 0.06 | 0.14 | 0.90 | 0.04 | 0.26 | 0.62 | 0.56 | 0.54 |

## 6. CONCLUSION

The paper proposes a method of convolutional networks training using additional data of the training sample such as level of distortion of input data. The experiment shows that the network trained by this way does not loss in quality, save its original architecture and obtains some additional properties such as the statistically dependence between network responses and level of distortion, as well as a significant error-free rate. In addition, the regression model of the dependence of confidence on the level of distortion was built to show a strong linear correlation in terms of medians. Nowadays many convolutional networks use only synthetic training data [14], which allows to create complex augmentation models and apply their parameters in training. The proposed method is very promising in terms of building reliable systems with a high level of confidence in responses.

## ACKNOWLEDGMENTS

The reported study was partially funded by RFBR according to the research projects 17-29-07093 and 18-07-01384.